\documentclass[preprint,10pt,nopreprintline]{elsarticle}
\usepackage[hidelinks]{hyperref}
\usepackage{iftex}
\ifPDFTeX
  \usepackage[T1]{fontenc}
  \usepackage[utf8]{inputenc}
  \usepackage{times}
  \usepackage{inconsolata}
\else
  \usepackage{fontspec}
  \IfFontExistsTF{Times New Roman}{
    \setmainfont{Times New Roman}
    \setmonofont{Times New Roman}
  }{
    \setmainfont{TeX Gyre Termes}
    \setmonofont{TeX Gyre Termes}
  }
\fi
\usepackage{latexsym}
\usepackage[nopatch=footnote]{microtype}
\usepackage{graphicx}
\usepackage{placeins}
\ifdefined\pdfobjcompresslevel\pdfobjcompresslevel=0\fi
\ifdefined\pdfminorversion\pdfminorversion=7\fi
\usepackage{booktabs}
\usepackage{longtable}
\usepackage{caption}
\usepackage{array}
\usepackage{amsmath}
\usepackage{amssymb}

\newcommand{\method}{LEAF-Mamba}
\newcommand{\lsr}{LEA}
\newcommand{\bgm}{EF}

\newcolumntype{L}[1]{>{\raggedright\arraybackslash}p{#1}}
\newcolumntype{C}[1]{>{\centering\arraybackslash}p{#1}}

\hypersetup{
    pdftitle={LEAF-Mamba: Feature Adaptation and Fusion for Whole Slide Image Analysis},
    pdfauthor={Lubin Gan; Jing Zhang; Mingjun Ma; Linhao Qu; Dachun Kai; Xiaoyan Sun},
    pdfsubject={Whole slide image analysis}
}
\usepackage[a4paper,top=4.3cm,bottom=4.3cm,left=4.8cm,right=4.8cm]{geometry}
\usepackage{setspace}
\usepackage{etoolbox}
\makeatletter
\patchcmd{\pprintMaketitle}{\Large\@title}{\fontsize{14}{17}\selectfont\@title}{}{\PackageError{Manuscript format}{Title font patch failed}{}}
\patchcmd{\pprintMaketitle}{\normalsize\elsauthors}{\fontsize{8}{10}\selectfont\elsauthors}{}{\PackageError{Manuscript format}{Author font patch failed}{}}
\patchcmd{\pprintMaketitle}{\vskip36pt}{\vskip6pt}{}{\PackageError{Manuscript format}{Frontmatter spacing patch failed}{}}
\patchcmd{\pprintMaketitle}{\vskip18pt}{\vskip10pt}{}{\PackageError{Manuscript format}{Title spacing patch failed}{}}
\patchcmd{\pprintMaketitle}{\par\vskip10pt}{\par\vskip6pt}{}{\PackageError{Manuscript format}{Author spacing patch failed}{}}
\patchcmd{\pprintMaketitle}{\hrule\vskip12pt}{\hrule\vskip8pt}{}{\PackageError{Manuscript format}{Abstract spacing patch failed}{}}
\patchcmd{\pprintMaketitle}{\thispagestyle{pprintTitle}}{\thispagestyle{plain}}{}{\PackageError{Manuscript format}{Title page number patch failed}{}}
\makeatother

\begin{document}

\begin{frontmatter}
\title{\method{}: Feature Adaptation and Fusion for Whole Slide Image Analysis}
\author[ustc,iai]{Lubin~Gan}
\ead{ganlubin@mail.ustc.edu.cn}
\author[ustc,iai]{Jing~Zhang\corref{cor}}
\ead{zhangjing@iai.ustc.edu.cn}
\author[zju]{Mingjun~Ma}
\ead{mingjun@tju.edu.cn}
\author[hms]{Linhao~Qu}
\ead{linhao_qu@hms.harvard.edu}
\author[ustc]{Dachun~Kai\fnref{present-dk}}
\ead{dachunkai@mail.ustc.edu.cn}
\author[ustc,iai]{Xiaoyan~Sun\corref{cor}}
\ead{sunxiaoyan@ustc.edu.cn}

\cortext[cor]{Corresponding authors: Jing Zhang and Xiaoyan Sun.}
\fntext[present-dk]{Present address: Nanyang Technological University, Singapore.}

\affiliation[ustc]{organization={National Engineering Laboratory for Brain-inspired Intelligence Technology and Application, MoE Key Laboratory of Brain-inspired Intelligent Perception and Cognition, University of Science and Technology of China},
    city={Hefei},
    postcode={230026},
    country={China}}
\affiliation[iai]{organization={Anhui Province Key Laboratory of Biomedical Imaging and Intelligent Processing, Institute of Artificial Intelligence, Hefei Comprehensive National Science Center},
    city={Hefei},
    postcode={230088},
    country={China}}
\affiliation[zju]{organization={School of Software Technology, Zhejiang University},
    city={Ningbo},
    postcode={315048},
    country={China}}
\affiliation[hms]{organization={Department of Biomedical Informatics, Harvard Medical School},
    city={Boston},
    state={MA},
    postcode={02115},
    country={USA}}

\begin{abstract}
Whole slide image (WSI) analysis enables computational diagnosis and prognosis, but gigapixel images and limited annotations make slide-level learning challenging. Multiple instance learning addresses this task by aggregating patch representations from frozen pretrained encoders. Recently, Mamba has enabled efficient modeling of long patch sequences. However, task supervision alone does not explicitly constrain information recovery during feature adaptation, while Mamba couples local processing with long-range propagation rather than retaining separate representations for fusion. Motivated by these challenges, we introduce \method{}, a framework that jointly addresses latent evidence adaptation and fusion for frozen-feature WSI analysis. A reconstruction-regularized adapter learns compact, task-specific representations while reconstructing the input features. Parallel Mamba and convolutional streams then capture long-range dependencies and sequence-local interactions, with an input-dependent gate modulating their joint representation before slide-level pooling. Experiments across feature extractors demonstrate improved average performance over strong MIL baselines in both diagnostic classification and survival prediction. Component analyses further support the contributions of adaptation and fusion. These findings highlight the value of jointly designing feature transformation and contextual aggregation to translate pretrained representations into effective slide-level predictions.

\end{abstract}
\end{frontmatter}

\section{Introduction}
\label{sec:introduction}

Whole slide images (WSIs) capture the cellular morphology and tissue organization used for computational cancer diagnosis and prognosis~\cite{srinidhi2021survey}.
Predicting a diagnostic category or patient outcome from a gigapixel slide requires aggregating evidence across heterogeneous tissue.
Dense region-level annotation demands extensive pathologist effort, so weakly supervised WSI models learn from slide- or patient-level labels~\cite{campanella2019clinical}.
Multiple instance learning (MIL) represents each slide as a bag of patches and learns a bag-level predictor.
Attention-based pooling weights patch contributions, while clustering-constrained learning adds a clustering objective to slide-level training~\cite{ilse2018attention,lu2021clam}.
Dual-stream methods identify critical instances, while feature-distillation methods learn slide representations through pseudo-bag training~\cite{li2021dsmil,zhang2022dtfd}.

In frozen-feature MIL, a pretrained encoder extracts patch embeddings, and a trainable aggregator maps them to slide-level predictions.
Pathology foundation models learn reusable patch representations from large tissue-image collections~\cite{chen2024foundation,vorontsov2024foundation}.
Transformer-based MIL models contextual interactions among patches~\cite{shao2021transmil}, while selective state-space models process long sequences with computational cost that scales linearly with sequence length~\cite{gu2024mamba}.
MambaMIL applies this sequence modeling to WSIs through patch reordering~\cite{yang2024mambamil}.
SemaMIL further studies semantic ordering and retrieval-guided state-space aggregation~\cite{gan2026semamil}, highlighting the role of sequence construction alongside the choice of sequence operator.
We focus on two stages of this pipeline: adapting frozen embeddings to the target task and combining short- and long-range contextual representations.

The first challenge is to adapt frozen features from bag-level supervision.
The frozen encoder receives no updates from the downstream diagnostic or survival loss, leaving feature adaptation to the slide-level model.
Re-embedding methods such as RRTMIL learn transformations before aggregation~\cite{tang2024rrtmil}.
The task loss supervises the transformed patches through the slide prediction, but does not explicitly constrain recovery of their input features.
A reconstruction loss supplies this missing constraint by penalizing input-recovery error for each compressed patch representation.
Joint optimization therefore gives the adapter two objectives: predict the slide-level outcome and reconstruct the frozen patch embeddings.

\begin{figure}[t]
\centering
\includegraphics[width=0.9\linewidth]{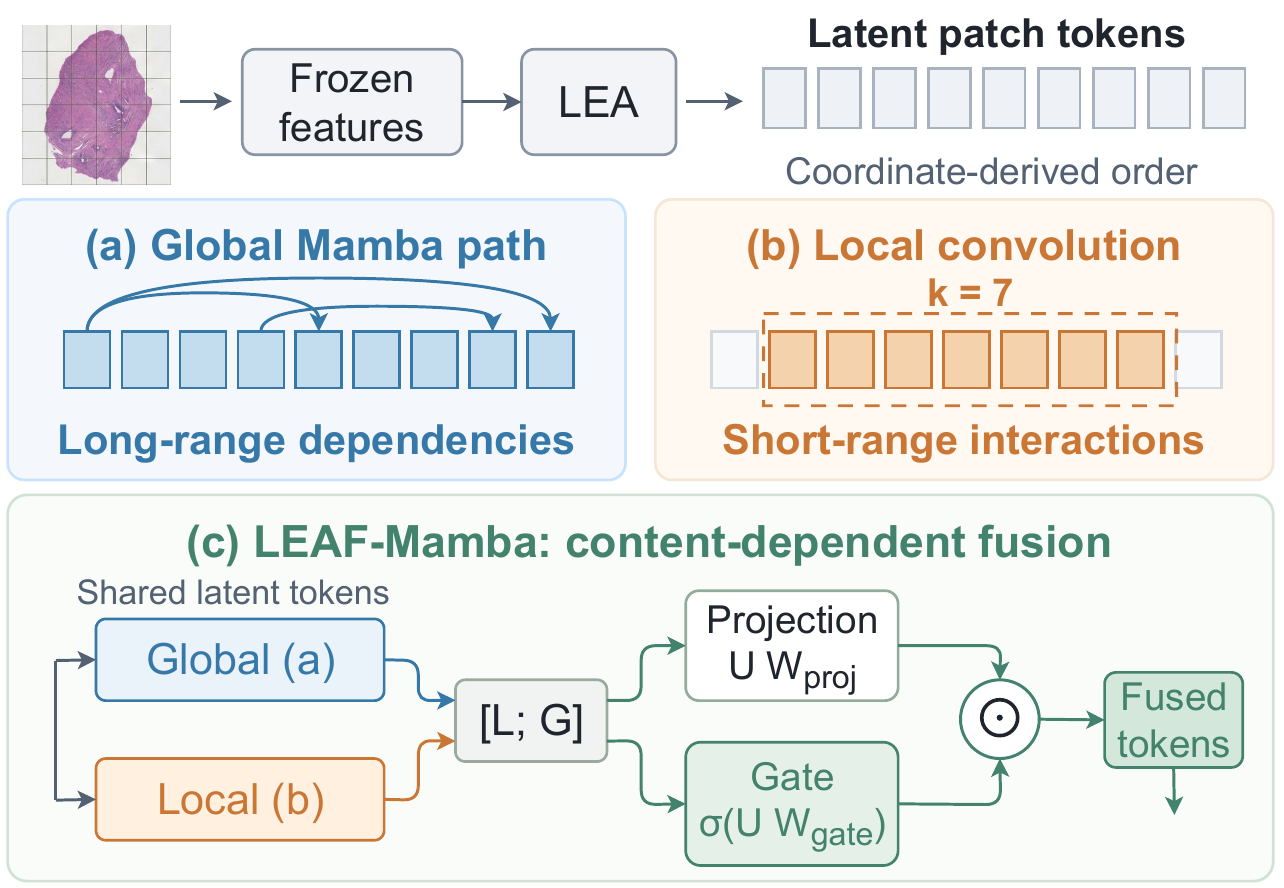}
\caption{Conceptual visualization of Mamba-guided evidence fusion.}
\label{fig:global_local_gated}
\end{figure}

The second challenge is to retain distinct short- and long-range representations for fusion.
In a standard Mamba block, local convolution feeds the selective state-space operation, placing short-range processing on the same path as long-range propagation~\cite{gu2024mamba}.
The convolutional output is not retained as a separate representation for fusion with the Mamba output.
A parallel convolutional stream preserves the sequence-local representation alongside the long-range output.
An input-dependent gate modulates their joint features before pooling.
Figure~\ref{fig:global_local_gated} illustrates the parallel streams and their input-dependent fusion.

We propose \method{}, a frozen-feature MIL framework that combines reconstruction-regularized latent evidence adaptation (\lsr{}) with Mamba-guided evidence fusion (\bgm{}). \lsr{} maps each patch embedding to a compact code through a nonlinear encoder and a linear skip projection.
The task loss and an auxiliary reconstruction loss jointly train the adapter, and the reconstruction decoder is removed at inference. \bgm{} processes the adapted tokens through parallel Mamba and convolutional streams along a fixed coordinate-derived ordering.
The convolutional stream models consecutive tokens in this sequence.
The two outputs are concatenated and projected, then modulated by a content-dependent sigmoid gate before slide-level pooling.

We evaluate \method{} on diagnostic classification and survival prediction using frozen ResNet-50, PLIP, and CONCH v1.5 features.
All reproduced baselines share the extracted features, data splits, and evaluation protocol. \method{} achieves the highest mean diagnostic AUC and mean survival C-Index among the compared methods with each encoder.
On the BRACS official split with CONCH v1.5 features, removing reconstruction reduces AUC from 76.8 to 75.8 under global-only aggregation.
With both branches enabled, input-dependent fusion improves AUC by 0.8 percentage points over a learned static gate.
Our main contributions are as follows:

\begin{itemize}
    \item We introduce a reconstruction-regularized latent adapter for frozen-feature MIL. Joint bag-level prediction and patchwise reconstruction train compact task-specific codes; the decoder is removed at inference.
    \item We develop an evidence-fusion module that retains parallel Mamba and sequence-local convolutional representations and modulates their projected joint features through an input-dependent gate before pooling.
    \item We show that \method{} achieves the highest mean diagnostic AUC and survival C-Index among the compared methods with each of three frozen encoders. Controlled ablations quantify the contributions of reconstruction and context fusion.
\end{itemize}

\section{Related Work}
\label{sec:related}

Related work for \method{} spans slide-level aggregation, long-sequence modeling, and frozen-feature adaptation.

\subsection{MIL for whole slide image analysis}
In weakly supervised WSI analysis, MIL represents a slide as a bag of patch features and learns from slide-level labels without dense annotation, despite large bag sizes and patch heterogeneity.
Attention-based MIL learns a slide summary from instance weights, and CLAM adds clustering constraints to make attention pooling more data-efficient in pathology~\cite{ilse2018attention,lu2021clam}.
DSMIL uses a dual stream to identify critical instances, while DTFD-MIL distills large bags through pseudo-bag training~\cite{li2021dsmil,zhang2022dtfd}.
Recent extensions shift the aggregation assumption in different directions: CAMIL adds neighborhood context, and ACMIL perturbs attention concentration through multi-branch masking~\cite{fourkioti2024camil,zhang2024acmil}.
RPMIL models bag predictions probabilistically, whereas Jigsaw learns order restoration from spatial correlations~\cite{zhao2025rpmil,chen2025jigsaw}.
DomSet-MIL addresses a different aspect of weak supervision: a dominant slide label can conceal coexisting lesions, motivating censored-label modeling and residual routing of lesion evidence~\cite{gan2026domsetmil}.
RRTMIL is the closest feature re-embedding comparator because it adapts pathology features for foundation-model transfer~\cite{tang2024rrtmil}.

\subsection{Long-sequence modeling in pathology}
Transformer and interventional MIL extend independent pooling through contextual dependency and confounder-aware bag learning~\cite{shao2021transmil,lin2023interventional}.
Graph-based models encode relational or spatial context through tissue graphs~\cite{zhao2020graph,liang2023graph}.
Hierarchical models aggregate local-to-slide representations across scales~\cite{chen2021patchgcn,chen2022hipt}.
Attention-based models can be computationally expensive for large bags, whereas graph and hierarchical models depend on relation construction or multi-scale design choices.
Structured state space and Mamba models offer a linear-time long-sequence alternative, with Mamba and VMamba establishing selective/state-space operators~\cite{gu2024mamba,liu2024vmamba}.
S4MIL and MambaMIL adapt this direction to WSI sequences~\cite{fillioux2023s4mil,yang2024mambamil}.
SemaMIL combines semantic reordering with retrieval-guided state-space modeling~\cite{gan2026semamil}; its sequence-construction strategy complements the study of parallel contextual streams.
MambaMIL is the closest sequence-side comparator to \method{}~\cite{yang2024mambamil}.

\subsection{Pathology foundation models and representation mismatch}
Pathology foundation models improve frozen patch representations: CTransPath uses pathology-specific contrastive pretraining, and PLIP adds vision-language supervision~\cite{wang2022ctranspath,huang2023plip}.
Broader general-purpose and clinical-grade encoders expand transfer across pathology tasks~\cite{chen2024foundation,vorontsov2024foundation}.
Recent multimodal whole-slide encoders extend this trend with image-text supervision~\cite{xu2024foundation,ding2025multimodal}.
FG-PAN refines visual patches and aligns them with fine-grained descriptions for zero-shot brain tumor subtyping~\cite{gan2026fgpan}. MeDKCoOp uses graph-structured visual and textual knowledge for biomedical prompt adaptation~\cite{wang2025medkcoop}. Both study how reusable vision-language representations can be matched to domain-specific evidence.
Frozen representations can still benefit from downstream task adaptation, as investigated by RRTMIL~\cite{tang2024rrtmil}. The feature mismatch addressed by this adaptation is distinct from generalization to a new acquisition domain.
The comparison of interest is how the adaptation objective and aggregation operators contribute under the same frozen-feature protocol.

\subsection{Adapters and hybrid sequence operators}
Bottleneck and residual adaptation have established precedents. Houlsby et al. insert trainable adapters within a frozen language model~\cite{houlsby2019adapters}, whereas CLIP-Adapter applies bottleneck transformations and residual blending to pretrained features~\cite{gao2024clipadapter}. These architectures motivate simple adapter controls; their supervision and insertion points differ from adaptation of offline WSI features. RRTMIL is the closer protocol-level reference because it re-embeds extracted patch features before MIL aggregation~\cite{tang2024rrtmil}. For aggregation, Conformer combines convolutional receptive fields with long-range attention in speech recognition~\cite{gulati2020conformer}. This supports a general design precedent for combining local and global operators, while the value of a chosen serialization in WSI analysis remains an empirical question.

Spatial and semantic constraints also arise in other medical imaging settings. SSCM combines spatial alignment and semantic consistency for multi-contrast MRI super-resolution~\cite{wu2026sscm}. Its cross-contrast reconstruction setting differs from recovering frozen patch features, but provides a related example of coordinating local detail with semantic representations. In vision-language modeling, geometry-aware dense-to-expert reconstruction studies the conversion of pretrained models into specialized expert modules~\cite{zhang2026geometrymoe}. Here, adaptation instead operates on offline patch embeddings before contextual aggregation.

\subsection{Structural information and distributed aggregation}
Preserving discriminative evidence during representation transformation is also studied in graph learning. GraphRefiner calibrates generic-token dominance in graph-text alignment~\cite{zhang2026graphrefiner}; G$^{2}$rammar encodes structural roles and semantic relationships~\cite{zheng2026g2rammar}. The H4G preprint studies how hyperbolic embedding geometry can retain fine-grained graph structure~\cite{zhang2025h4g}. These formulations provide related perspectives on distinguishing useful local information from a broad contextual representation.

Multi-agent research examines a complementary question: how should distributed information sources communicate? HyperAgent uses hyperedges to represent collaboration groups~\cite{zhang2026hyperagent}, while EvoHyper evolves a shared hypergraph of agents and memory~\cite{zhang2026evohyper}. D$^{3}$MAS coordinates task decomposition, reasoning, and distributed memory~\cite{zhang2026d3mas}, and graph-decoupling methods study knowledge sharing among agents~\cite{zhang2026graphdecoupling}. At a conceptual level, these works connect the organization of individual information sources with their collective use. In WSI analysis, the corresponding design problem concerns patch representations and their aggregation under slide-level supervision.

\section{Method}
\label{sec:method}

\begin{figure}[!t]
\centering
\includegraphics[width=\textwidth]{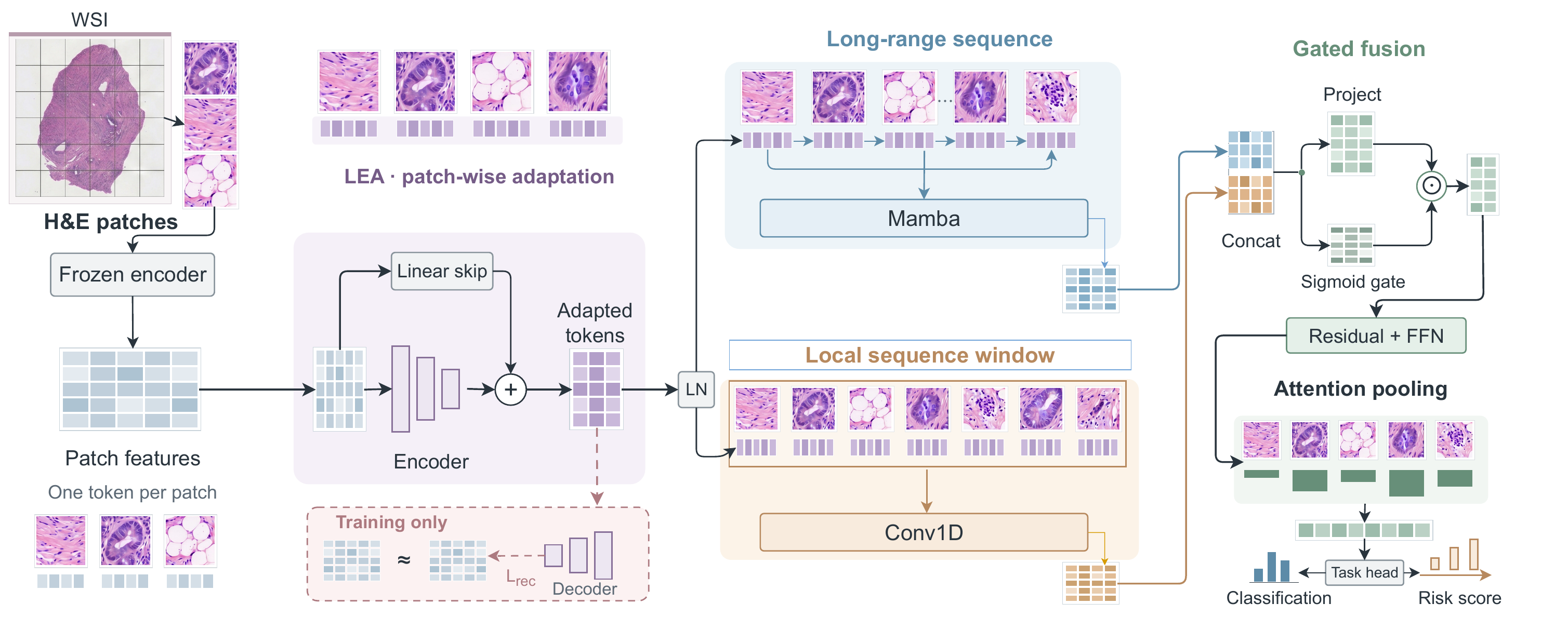}
\caption{Overview of \method{}. Frozen features undergo reconstruction-regularized latent evidence adaptation and Mamba-guided fusion of long-range sequence context, short-range token interactions, and content-dependent gating for slide-level prediction.}
\label{fig:framework}
\end{figure}

\method{} transforms frozen patch embeddings, models them through long-range and sequence-local paths, and combines the resulting features before gated attention pooling. The adapter and aggregator are trained jointly with slide-level supervision, while the patch encoder remains frozen.
Figure~\ref{fig:framework} summarizes the complete pipeline.

\subsection{Problem Setup}

Following the standard MIL paradigm, the $i$-th WSI is represented as a bag
\begin{equation}
    \mathcal{B}_i = \{x_{ij}\}_{j=1}^{N_i},
\end{equation}
where $N_i$ is the number of patches and $x_{ij}$ denotes a raw image patch. A frozen feature extractor maps each patch to $h_{ij} \in \mathbb{R}^{D}$, producing a bag feature matrix $H_i \in \mathbb{R}^{N_i \times D}$. We denote the slide-level target by $\mathcal{Y}_i$: for diagnostic classification, $\mathcal{Y}_i=y_i$ with $y_i \in \{0,1,\ldots,C-1\}$; for survival prediction, $\mathcal{Y}_i=(t_i,\delta_i)$ contains the observed time and event indicator. The goal is to learn a mapping $F(H_i)$ that produces class logits or a survival risk score from frozen bag features.

\subsection{Latent Evidence Adaptation}

\begin{figure}[!t]
\centering
\includegraphics[width=\linewidth]{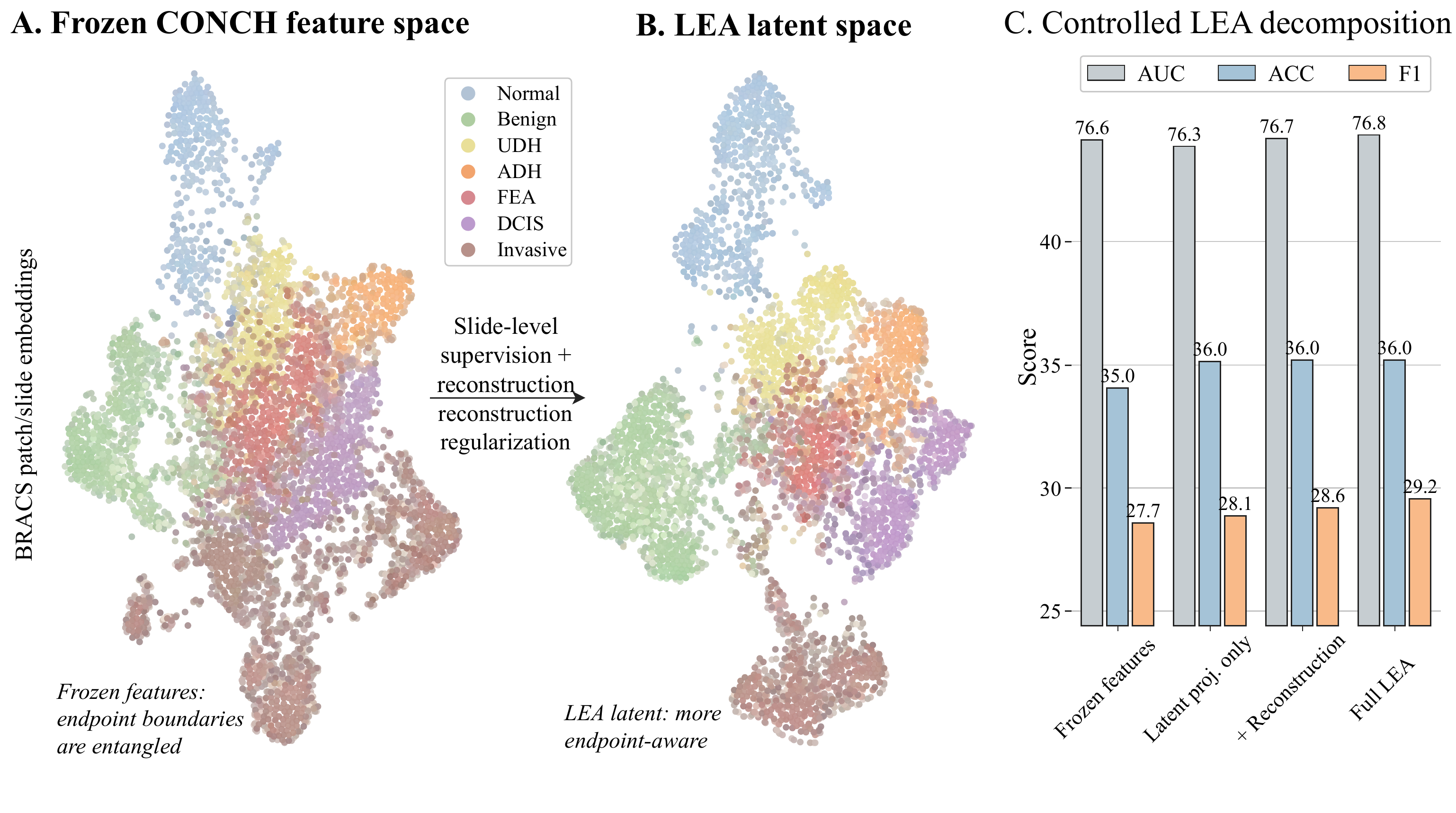}
\caption{Effect of \lsr{} on BRACS features.}
\label{fig:lea_embedding}
\end{figure}

\lsr{} formulates task-specific feature adaptation with an auxiliary reconstruction objective before aggregation.
The input is the frozen feature matrix $H_i$, whose representation was learned before the downstream MIL task.
Slide-level supervision trains the adapter jointly with the aggregator; reconstructing $H_i$ adds an input-recovery objective to the task loss.
\lsr{} projects each token into a lower-dimensional code from which a decoder reconstructs the original feature.
The module contains a nonlinear encoder $E(\cdot)$, a decoder $G(\cdot)$, and a linear skip projection $P_{\mathrm{skip}}(\cdot)$.
We define the latent representation as
\begin{equation}
    Z_i = E(H_i) + P_{\mathrm{skip}}(H_i) \in \mathbb{R}^{N_i \times d}.
\end{equation}
The encoder and decoder are lightweight projection heads applied token-wise, with
\begin{equation}
    E(h) = W^E_2 \phi(W^E_1 h), \quad G(z) = W^G_2 \phi(W^G_1 z),
\end{equation}
where $\phi(\cdot)$ is GELU and the latent width satisfies $d < D$. The nonlinear encoder learns the task-adaptive transformation, while the linear skip projection supplies a second route from the frozen input. The decoder reconstructs the original feature matrix:
\begin{equation}
    \hat{H}_i = G(Z_i) \in \mathbb{R}^{N_i \times D}.
\end{equation}
The reconstruction loss is
\begin{equation}
    \mathcal{L}_{\mathrm{rec}} =
    \frac{1}{N_i}\sum_{j=1}^{N_i}\left\|h_{ij} - \hat{h}_{ij}\right\|_2^2.
\end{equation}
The overall objective is
\begin{equation}
    \mathcal{L} = \mathcal{L}_{\mathrm{task}} + \lambda_{\mathrm{rec}} \mathcal{L}_{\mathrm{rec}},
\end{equation}
where $\mathcal{L}_{\mathrm{task}}$ is cross-entropy for diagnostic classification and the negative Cox partial log-likelihood for survival prediction. The task loss trains predictive features, and the reconstruction term penalizes input-recovery error. Figure~\ref{fig:lea_embedding} compares the distributions of frozen CONCH features and task-trained latent features on BRACS, alongside predictive performance for the projection, reconstruction, and skip-connection configurations.

\subsection{Global-Local Evidence Modeling}

\bgm{} applies two sequence operators before bag-level pooling.
One stream models long-range dependencies, while the other captures interactions within a finite convolutional window. Combining these receptive fields follows the hybrid-operator design principle used in Conformer~\cite{gulati2020conformer}.
For coordinate-bearing bags, raster serialization sorts patch coordinates by $y$ first and then $x$, preserving input order for exact ties. The ordered sequence is retained across blocks. Let $Z_i^{(0)}=Z_i$ denote the input to the first evidence-fusion block. At block $l$, we normalize the current token sequence:
\begin{equation}
    \tilde{Z}_i^{(l-1)} = \mathrm{LN}(Z_i^{(l-1)}).
\end{equation}
The global stream uses a selective state space operation $\Phi_{\mathrm{SSM}}(\cdot)$ to model long-range dependencies:
\begin{equation}
    Z_{\mathrm{global}}^{(l)} = \Phi_{\mathrm{SSM}}(\tilde{Z}_i^{(l-1)}).
\end{equation}
Because Mamba-style models are sequence-order sensitive, we use the same coordinate-derived ordering for \method{} and MambaMIL in all comparisons. Alongside the global stream, a separate local stream models short-range token interactions using depthwise separable convolutions:
\begin{equation}
    Z_{\mathrm{local}}^{(l)} =
    \phi\left(\mathrm{PWConv}\left(\mathrm{DWConv}_{k}(\tilde{Z}_i^{(l-1)})\right)\right),
\end{equation}
where $\mathrm{DWConv}_{k}$ is a depthwise one-dimensional convolution with kernel size $k$, $\mathrm{PWConv}$ is a pointwise convolution, and $\phi(\cdot)$ is the GELU activation. The default kernel spans seven consecutive tokens in the raster-ordered sequence, defining a \emph{sequence-local receptive field}. Convolutional windows follow this sequence across row transitions, tissue holes, and disconnected regions. The kernel size determines the number of neighboring tokens processed at each position.

\subsection{Content-Dependent Evidence Fusion}

Gated fusion combines the two evidence streams through input-dependent modulation of their joint representation.
We first concatenate the two streams into
$U_i^{(l)}=[Z_{\mathrm{local}}^{(l)};Z_{\mathrm{global}}^{(l)}]\in \mathbb{R}^{N_i \times 2d}$ and apply a gated projection:
\begin{equation}
    Z_{\mathrm{fuse}}^{(l)} =
    (U_i^{(l)} W_{\mathrm{proj}}) \odot \sigma(U_i^{(l)} W_{\mathrm{gate}}),
\end{equation}
where $W_{\mathrm{proj}}, W_{\mathrm{gate}} \in \mathbb{R}^{2d \times d}$ are learnable parameters, $\sigma(\cdot)$ is the sigmoid function, and $\odot$ denotes element-wise multiplication. The sigmoid gate modulates the projected joint representation, allowing the fused feature to respond differently across slides and tokens. A residual feed-forward update then refines the representation:
\begin{equation}
    Z_i^{(l)} = Z_i^{(l-1)} +
    M\left(\mathrm{LN}\left(Z_i^{(l-1)} + Z_{\mathrm{fuse}}^{(l)}\right)\right),
\end{equation}
where $Z_i^{(l-1)}$ and $Z_i^{(l)}$ denote the block input and output, $M(\cdot)$ is an MLP, and $\mathrm{LN}(\cdot)$ denotes layer normalization. Gated attention pools the final normalized token sequence into a bag representation, which is passed to the slide-level task head. Figure~\ref{fig:global_local_gated} summarizes the long-range stream, sequence-local stream, and input-dependent feature modulation.

\section{Experiments}
\label{sec:experiments}

We organize the experiments around three questions.
First, does the representation-and-evidence design remain useful across feature extractors?
Second, does the same principle carry from diagnostic classification to noisier survival prediction?
Third, do controlled variants support the roles assigned to \lsr{}, local modeling, and gated fusion?
We first define the datasets and unified protocol, then answer these questions through benchmark comparisons and targeted ablations.

\subsection{Datasets and Protocol}

We evaluate \method{} on both diagnostic classification and survival prediction under a unified frozen-feature protocol. For diagnostic classification, we use EBRAINS for 30-class brain tumor subtyping~\cite{roetzer2022digital}, BRACS for 7-class breast lesion classification~\cite{brancati2022bracs}, and Camelyon16 for breast cancer metastasis detection~\cite{camelyon2017}. BRACS is evaluated with both 5-fold cross-validation and its official split. For survival prediction, we use five cohorts from The Cancer Genome Atlas (TCGA): BLCA, BRCA, COADREAD, STAD, and HNSC~\cite{weinstein2013tcga}. Prognostic performance is measured by Concordance Index (C-Index) and reported as C-Index $\times 100$.

\subsection{Implementation Details}

All reproduced baselines reuse the same pre-extracted features, split protocol, early-stopping rule, and metric computation as \method{}.
For Mamba-based models, the same fixed coordinate-derived patch serialization is used throughout so that ordering does not confound the comparison.
This protocol isolates the slide-level aggregator stage under frozen features.

\method{} uses two evidence-fusion blocks unless otherwise stated.
In the \lsr{} module, the encoder maps the input dimension $D$ to a latent width $d$ through a hidden width of $2d$, uses layer normalization, GELU, and dropout $0.25$, and adds a linear skip projection from the frozen feature.
The decoder mirrors this projection to reconstruct the original frozen feature and contributes the mean-squared reconstruction loss during training.
Each evidence-fusion block applies a local depthwise one-dimensional convolution with kernel size $7$, followed by a pointwise convolution, and a global Mamba path with state dimension $32$, convolution width $4$, and expansion factor $2$.
The local and global outputs are concatenated, projected, and modulated by a sigmoid fusion gate before a residual feed-forward update.
After the final block, gated attention pools the refined token features into a slide representation, which is passed to a linear task head.

\subsection{Comparison with Recent WSI Baselines}

\begin{figure}[!t]
\centering
\includegraphics[width=\textwidth]{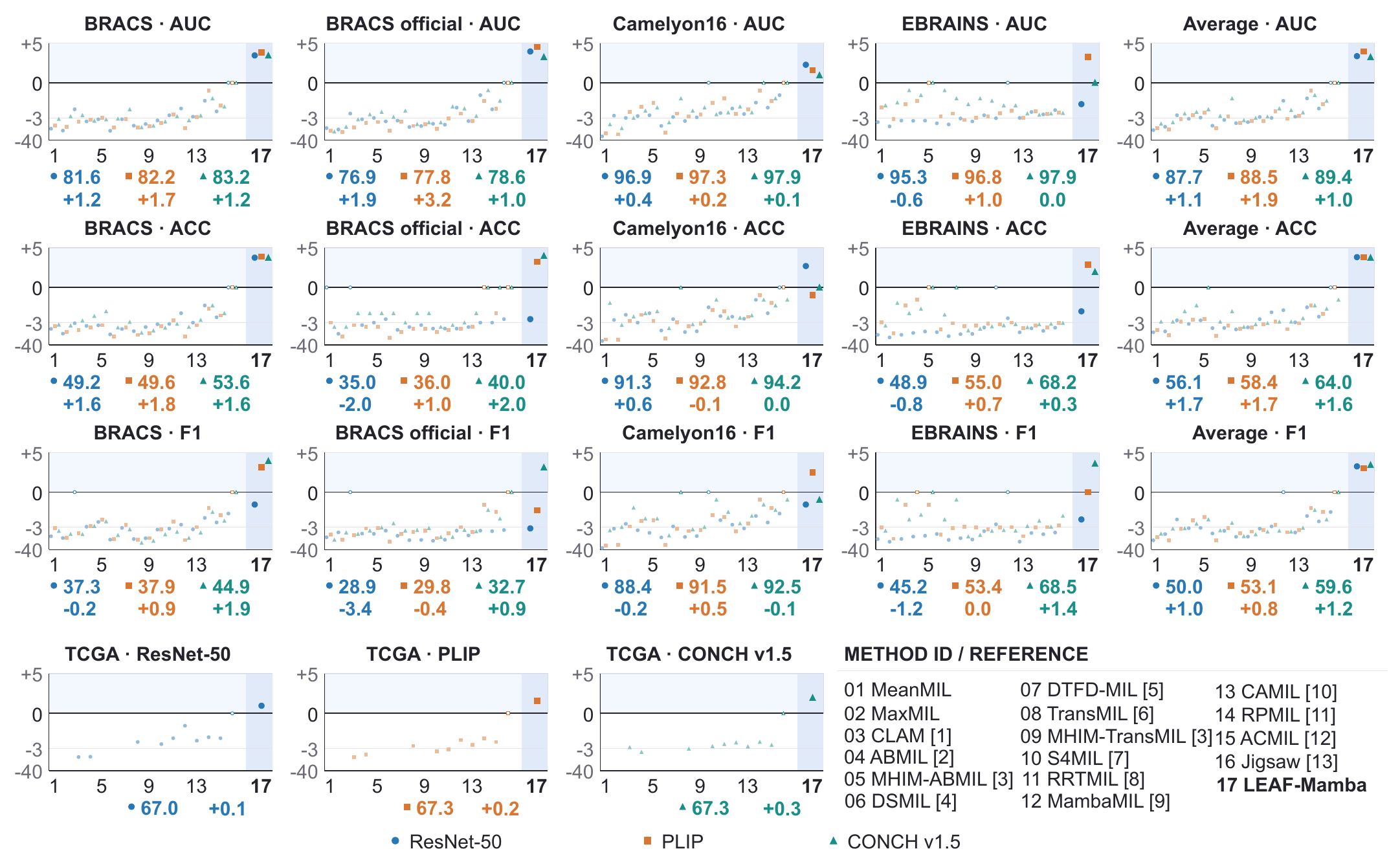}
\caption{Diagnostic classification and TCGA survival prediction results across frozen feature extractors. The vertical axis shows the score difference from the best baseline using the same feature extractor, on a nonlinear scale.}
\label{fig:main_results}
\end{figure}

To present the extensive comparisons within the available space, Figure~\ref{fig:main_results} uses a compact multi-panel layout showing score differences relative to the strongest baseline for each feature extractor.
Tables~\ref{tab:diagnostic_resnet}--\ref{tab:survival} report the complete numerical results corresponding to Figure~\ref{fig:main_results}.

Figure~\ref{fig:main_results} includes recent WSI baselines.
The MHIM variants combine masked hard-instance mining with ABMIL or TransMIL backbones~\cite{tang2023mhim}.
CAMIL and RPMIL provide context-aware and probabilistic WSI comparators, respectively~\cite{fourkioti2024camil,zhao2025rpmil}.
ACMIL and the instance-jigsaw model provide attention-challenging and order-restoration comparators~\cite{zhang2024acmil,chen2025jigsaw}.

In Figure~\ref{fig:main_results}, \method{} achieves strongest average diagnostic performance for each feature extractor under shared frozen-feature protocol.

Specifically, average AUC reaches 87.7, 88.5, and 89.4 with ResNet-50, PLIP, and CONCH v1.5 features, respectively, exceeding the strongest baseline, Jigsaw, by 1.1, 1.9, and 1.0 percentage points. This consistent ranking shows benefits across all three frozen feature extractors. The BRACS official split provides a representative benchmark-level example: with PLIP features, \method{} achieves AUC 77.8, exceeding Jigsaw's 74.6 by 3.2 percentage points. Together, the average results and this task-specific comparison support the effectiveness of combining feature adaptation with sequence aggregation for diagnostic classification.

Beyond diagnostic classification, Figure~\ref{fig:main_results} shows that \method{} also achieves the best average C-Index with each evaluated feature extractor.

Across the five TCGA cohorts, the average C-Index $\times 100$ is 67.0 with ResNet-50 features and 67.3 with PLIP and CONCH v1.5 features. These scores exceed Jigsaw, the strongest survival baseline, by 0.1, 0.2, and 0.3 points, respectively. The leading average performance across all three feature extractors complements the diagnostic results, showing that the combined adaptation and aggregation design is effective for both slide-level classification and survival prediction. This design therefore supports both prediction tasks evaluated here.

\FloatBarrier

\begingroup
\setstretch{1}
\fontsize{10}{12}\selectfont
\setlength{\tabcolsep}{3pt}
\renewcommand{\arraystretch}{1.0}
\setlength{\LTcapwidth}{\linewidth}
\setlength{\LTpre}{10pt}
\setlength{\LTpost}{12pt}
\begin{longtable}{@{}p{\linewidth}@{}}
\caption{Diagnostic classification results with ResNet-50 features. All scores are percentages; higher is better.}\label{tab:diagnostic_resnet}\\
\endfirsthead
\caption[]{Diagnostic classification results with ResNet-50 features. All scores are percentages; higher is better (continued).}\\
\endhead
\endfoot
\endlastfoot
\begin{minipage}{\linewidth}
\fontsize{10}{12}\selectfont
\textbf{(a) BRACS and BRACS official}\par\smallskip
\begin{tabular*}{\linewidth}{@{\extracolsep{\fill}}lrrrrrr@{}}
\toprule
& \multicolumn{3}{c}{BRACS} & \multicolumn{3}{c}{BRACS official} \\
\cmidrule(lr){2-4}\cmidrule(lr){5-7}
Method & AUC & ACC & F1 & AUC & ACC & F1 \\
\midrule
MeanMIL & 69.9 & 41.4 & 29.0 & 65.2 & \textbf{37.0} & 22.8 \\ 
MaxMIL & 67.3 & 36.8 & 27.1 & 63.5 & 26.0 & 26.9 \\ 
CLAM & 79.3 & 44.4 & \textbf{37.5} & 73.2 & \textbf{37.0} & \textbf{32.3} \\ 
ABMIL & 75.9 & 43.2 & 34.7 & 71.5 & 33.0 & 26.1 \\ 
MHIM-ABMIL & 76.9 & 43.5 & 35.9 & 72.0 & 34.0 & 25.9 \\ 
DSMIL & 67.0 & 35.8 & 24.9 & 70.5 & 35.0 & 27.9 \\ 
DTFD-MIL & 76.9 & 41.3 & 34.3 & 70.9 & 30.0 & 24.3 \\ 
TransMIL & 73.0 & 39.5 & 25.8 & 66.4 & 32.0 & 18.9 \\ 
MHIM-TransMIL & 74.3 & 42.6 & 26.8 & 67.3 & 33.0 & 19.3 \\ 
S4MIL & 76.4 & 44.0 & 33.6 & 69.1 & 31.0 & 18.0 \\ 
RRTMIL & 78.2 & 44.1 & 34.0 & 68.4 & 31.0 & 19.6 \\ 
MambaMIL & 79.4 & 46.6 & 35.3 & 74.1 & 31.0 & 27.9 \\ 
CAMIL & 77.8 & 44.6 & 33.7 & 70.7 & 32.0 & 24.6 \\ 
RPMIL & 80.0 & 47.2 & 36.5 & 74.8 & 34.0 & 27.9 \\ 
ACMIL & 79.0 & 46.0 & 35.8 & 73.9 & 34.0 & 27.6 \\ 
Jigsaw & 80.4 & 47.6 & 36.9 & 75.0 & 35.0 & 28.2 \\ 
LEAF-Mamba & \textbf{81.6} & \textbf{49.2} & 37.3 & \textbf{76.9} & 35.0 & 28.9 \\ 
\bottomrule
\end{tabular*}
\par\smallskip{\footnotesize Best scores in each column are in bold.\par}
\end{minipage} \\
\noalign{\vskip 10pt}
\begin{minipage}{\linewidth}
\fontsize{10}{12}\selectfont
\textbf{(b) Camelyon16 and EBRAINS}\par\smallskip
\begin{tabular*}{\linewidth}{@{\extracolsep{\fill}}lrrrrrr@{}}
\toprule
& \multicolumn{3}{c}{Camelyon16} & \multicolumn{3}{c}{EBRAINS} \\
\cmidrule(lr){2-4}\cmidrule(lr){5-7}
Method & AUC & ACC & F1 & AUC & ACC & F1 \\
\midrule
MeanMIL & 70.9 & 65.2 & 55.4 & 91.0 & 37.3 & 34.6 \\ 
MaxMIL & 94.2 & 88.5 & 85.1 & 87.9 & 32.9 & 30.4 \\ 
CLAM & 93.5 & 89.5 & 86.6 & 91.8 & 37.3 & 34.8 \\ 
ABMIL & 94.1 & 87.7 & 84.4 & 91.7 & 39.7 & 37.6 \\ 
MHIM-ABMIL & 94.3 & 89.7 & 82.9 & 91.8 & 41.3 & 38.4 \\ 
DSMIL & 88.9 & 83.2 & 78.9 & 90.3 & 40.6 & 39.9 \\ 
DTFD-MIL & 93.8 & 88.7 & 85.9 & 89.5 & 35.2 & 33.5 \\ 
TransMIL & 92.2 & 81.7 & 79.4 & 92.7 & 38.2 & 35.8 \\ 
MHIM-TransMIL & 93.7 & 82.5 & 80.3 & 91.4 & 38.6 & 34.0 \\ 
S4MIL & 96.5 & 90.2 & \textbf{88.6} & 93.7 & 44.4 & 41.7 \\ 
RRTMIL & 94.8 & 89.3 & 87.3 & 92.8 & \textbf{49.7} & 44.3 \\ 
MambaMIL & 95.5 & 89.2 & 86.4 & \textbf{95.9} & 47.9 & \textbf{46.4} \\ 
CAMIL & 94.6 & 88.8 & 85.2 & 92.9 & 40.4 & 30.5 \\ 
RPMIL & 96.0 & 90.5 & 87.8 & 94.4 & 43.9 & 40.6 \\ 
ACMIL & 95.6 & 89.7 & 86.9 & 93.9 & 41.9 & 39.7 \\ 
Jigsaw & 96.3 & 90.7 & 88.0 & 94.7 & 44.3 & 40.9 \\ 
LEAF-Mamba & \textbf{96.9} & \textbf{91.3} & 88.4 & 95.3 & 48.9 & 45.2 \\ 
\bottomrule
\end{tabular*}
\par\smallskip{\footnotesize Best scores in each column are in bold.\par}
\end{minipage} \\
\noalign{\vskip 10pt}
\begin{minipage}{\linewidth}
\fontsize{10}{12}\selectfont
\textbf{(c) Average}\par\smallskip
\begin{tabular*}{\linewidth}{@{\extracolsep{\fill}}lrrr@{}}
\toprule
Method & AUC & ACC & F1 \\
\midrule
MeanMIL & 74.2 & 45.2 & 35.4 \\ 
MaxMIL & 78.2 & 46.1 & 42.4 \\ 
CLAM & 84.4 & 52.1 & 47.8 \\ 
ABMIL & 83.3 & 50.9 & 45.7 \\ 
MHIM-ABMIL & 83.7 & 52.1 & 45.8 \\ 
DSMIL & 79.2 & 48.7 & 42.9 \\ 
DTFD-MIL & 82.8 & 48.8 & 44.5 \\ 
TransMIL & 81.1 & 47.8 & 40.0 \\ 
MHIM-TransMIL & 81.7 & 49.2 & 40.1 \\ 
S4MIL & 83.9 & 52.4 & 45.5 \\ 
RRTMIL & 83.6 & 53.5 & 46.3 \\ 
MambaMIL & 86.2 & 53.7 & 49.0 \\ 
CAMIL & 84.0 & 51.5 & 43.5 \\ 
RPMIL & 86.3 & 53.9 & 48.2 \\ 
ACMIL & 85.6 & 52.9 & 47.5 \\ 
Jigsaw & 86.6 & 54.4 & 48.5 \\ 
LEAF-Mamba & \textbf{87.7} & \textbf{56.1} & \textbf{50.0} \\ 
\bottomrule
\end{tabular*}
\par\smallskip{\footnotesize Best scores in each column are in bold. Average summarizes the four diagnostic settings.\par}
\end{minipage} \\
\end{longtable}
\endgroup

\begingroup
\setstretch{1}
\fontsize{10}{12}\selectfont
\setlength{\tabcolsep}{3pt}
\renewcommand{\arraystretch}{1.0}
\setlength{\LTcapwidth}{\linewidth}
\setlength{\LTpre}{10pt}
\setlength{\LTpost}{12pt}
\begin{longtable}{@{}p{\linewidth}@{}}
\caption{Diagnostic classification results with PLIP features. All scores are percentages; higher is better.}\label{tab:diagnostic_plip}\\
\endfirsthead
\caption[]{Diagnostic classification results with PLIP features. All scores are percentages; higher is better (continued).}\\
\endhead
\endfoot
\endlastfoot
\begin{minipage}{\linewidth}
\fontsize{10}{12}\selectfont
\textbf{(a) BRACS and BRACS official}\par\smallskip
\begin{tabular*}{\linewidth}{@{\extracolsep{\fill}}lrrrrrr@{}}
\toprule
& \multicolumn{3}{c}{BRACS} & \multicolumn{3}{c}{BRACS official} \\
\cmidrule(lr){2-4}\cmidrule(lr){5-7}
Method & AUC & ACC & F1 & AUC & ACC & F1 \\
\midrule
MeanMIL & 73.2 & 43.3 & 33.7 & 61.2 & 32.0 & 23.2 \\ 
MaxMIL & 72.1 & 38.7 & 26.8 & 59.0 & 20.0 & 17.8 \\ 
CLAM & 75.9 & 40.7 & 31.0 & 65.6 & 31.0 & 22.9 \\ 
ABMIL & 76.9 & 42.5 & 34.4 & 69.1 & 30.0 & 26.8 \\ 
MHIM-ABMIL & 77.6 & 43.8 & 35.8 & 69.8 & 30.0 & 26.9 \\ 
DSMIL & 71.4 & 33.1 & 25.1 & 61.2 & 17.0 & 14.4 \\ 
DTFD-MIL & 77.1 & 42.5 & 34.7 & 70.1 & 26.0 & 23.8 \\ 
TransMIL & 71.2 & 35.3 & 24.0 & 67.1 & 29.0 & 25.1 \\ 
MHIM-TransMIL & 72.5 & 37.5 & 26.5 & 68.2 & 29.0 & 25.8 \\ 
S4MIL & 75.2 & 43.3 & 33.1 & 64.3 & 26.0 & 19.7 \\ 
RRTMIL & 78.5 & 45.5 & 35.5 & 71.5 & 28.0 & 24.2 \\ 
MambaMIL & 70.5 & 33.0 & 23.8 & 72.8 & 29.0 & 27.3 \\ 
CAMIL & 77.8 & 43.8 & 34.1 & 70.2 & 30.0 & 25.0 \\ 
RPMIL & 80.4 & 47.2 & 36.7 & 74.2 & 35.0 & 30.0 \\ 
ACMIL & 79.8 & 46.5 & 36.3 & 73.6 & 34.0 & 29.7 \\ 
Jigsaw & 80.5 & 47.8 & 37.0 & 74.6 & 35.0 & \textbf{30.2} \\ 
LEAF-Mamba & \textbf{82.2} & \textbf{49.6} & \textbf{37.9} & \textbf{77.8} & \textbf{36.0} & 29.8 \\ 
\bottomrule
\end{tabular*}
\par\smallskip{\footnotesize Best scores in each column are in bold.\par}
\end{minipage} \\
\noalign{\vskip 10pt}
\begin{minipage}{\linewidth}
\fontsize{10}{12}\selectfont
\textbf{(b) Camelyon16 and EBRAINS}\par\smallskip
\begin{tabular*}{\linewidth}{@{\extracolsep{\fill}}lrrrrrr@{}}
\toprule
& \multicolumn{3}{c}{Camelyon16} & \multicolumn{3}{c}{EBRAINS} \\
\cmidrule(lr){2-4}\cmidrule(lr){5-7}
Method & AUC & ACC & F1 & AUC & ACC & F1 \\
\midrule
MeanMIL & 78.7 & 71.9 & 65.4 & 94.9 & 51.5 & 50.3 \\ 
MaxMIL & 77.1 & 71.0 & 67.3 & 92.3 & 43.9 & 42.1 \\ 
CLAM & 95.6 & 90.5 & 88.3 & 95.2 & 53.9 & 53.2 \\ 
ABMIL & 96.3 & 91.7 & 89.6 & 95.4 & 54.1 & \textbf{53.4} \\ 
MHIM-ABMIL & 96.6 & 90.5 & 89.7 & 95.8 & 54.3 & 53.2 \\ 
DSMIL & 80.5 & 73.9 & 67.3 & 93.3 & 47.7 & 46.3 \\ 
DTFD-MIL & 95.3 & 90.0 & 87.4 & 94.0 & 51.1 & 50.1 \\ 
TransMIL & 94.5 & 88.5 & 86.4 & 92.0 & 49.9 & 50.6 \\ 
MHIM-TransMIL & 95.1 & 89.5 & 86.9 & 92.8 & 50.2 & 50.4 \\ 
S4MIL & 96.6 & 92.2 & 90.3 & 93.2 & 50.9 & 49.5 \\ 
RRTMIL & 96.8 & 92.1 & 90.1 & 94.1 & 50.7 & 50.3 \\ 
MambaMIL & 95.2 & 88.2 & 86.2 & 94.5 & 51.3 & 50.5 \\ 
CAMIL & 96.0 & 91.4 & 89.1 & 92.8 & 48.6 & 45.0 \\ 
RPMIL & 97.0 & 92.8 & 90.9 & 94.4 & 50.0 & 50.4 \\ 
ACMIL & 96.7 & 92.6 & 90.7 & 93.9 & 49.1 & 50.5 \\ 
Jigsaw & 97.1 & \textbf{92.9} & 91.0 & 94.2 & 51.2 & 51.0 \\ 
LEAF-Mamba & \textbf{97.3} & 92.8 & \textbf{91.5} & \textbf{96.8} & \textbf{55.0} & \textbf{53.4} \\ 
\bottomrule
\end{tabular*}
\par\smallskip{\footnotesize Best scores in each column are in bold.\par}
\end{minipage} \\
\noalign{\vskip 10pt}
\begin{minipage}{\linewidth}
\fontsize{10}{12}\selectfont
\textbf{(c) Average}\par\smallskip
\begin{tabular*}{\linewidth}{@{\extracolsep{\fill}}lrrr@{}}
\toprule
Method & AUC & ACC & F1 \\
\midrule
MeanMIL & 77.0 & 49.7 & 43.1 \\ 
MaxMIL & 75.1 & 43.4 & 38.5 \\ 
CLAM & 83.1 & 54.0 & 48.8 \\ 
ABMIL & 84.4 & 54.6 & 51.0 \\ 
MHIM-ABMIL & 85.0 & 54.7 & 51.4 \\ 
DSMIL & 76.6 & 42.9 & 38.3 \\ 
DTFD-MIL & 84.1 & 52.4 & 49.0 \\ 
TransMIL & 81.2 & 50.7 & 46.5 \\ 
MHIM-TransMIL & 82.2 & 51.6 & 47.4 \\ 
S4MIL & 82.3 & 53.1 & 48.1 \\ 
RRTMIL & 85.2 & 54.1 & 50.0 \\ 
MambaMIL & 83.3 & 50.4 & 47.0 \\ 
CAMIL & 84.2 & 53.5 & 48.3 \\ 
RPMIL & 86.5 & 56.3 & 52.0 \\ 
ACMIL & 86.0 & 55.6 & 51.8 \\ 
Jigsaw & 86.6 & 56.7 & 52.3 \\ 
LEAF-Mamba & \textbf{88.5} & \textbf{58.4} & \textbf{53.1} \\ 
\bottomrule
\end{tabular*}
\par\smallskip{\footnotesize Best scores in each column are in bold. Average summarizes the four diagnostic settings.\par}
\end{minipage} \\
\end{longtable}
\endgroup

\begingroup
\setstretch{1}
\fontsize{10}{12}\selectfont
\setlength{\tabcolsep}{3pt}
\renewcommand{\arraystretch}{1.0}
\setlength{\LTcapwidth}{\linewidth}
\setlength{\LTpre}{10pt}
\setlength{\LTpost}{12pt}
\begin{longtable}{@{}p{\linewidth}@{}}
\caption{Diagnostic classification results with CONCH v1.5 features. All scores are percentages; higher is better.}\label{tab:diagnostic_conch}\\
\endfirsthead
\caption[]{Diagnostic classification results with CONCH v1.5 features. All scores are percentages; higher is better (continued).}\\
\endhead
\endfoot
\endlastfoot
\begin{minipage}{\linewidth}
\fontsize{10}{12}\selectfont
\textbf{(a) BRACS and BRACS official}\par\smallskip
\begin{tabular*}{\linewidth}{@{\extracolsep{\fill}}lrrrrrr@{}}
\toprule
& \multicolumn{3}{c}{BRACS} & \multicolumn{3}{c}{BRACS official} \\
\cmidrule(lr){2-4}\cmidrule(lr){5-7}
Method & AUC & ACC & F1 & AUC & ACC & F1 \\
\midrule
MeanMIL & 78.6 & 48.2 & 38.4 & 62.8 & 33.0 & 19.1 \\ 
MaxMIL & 79.5 & 47.8 & 36.4 & 69.2 & 33.0 & 26.3 \\ 
CLAM & 79.8 & 49.6 & 36.3 & 74.0 & 37.0 & 25.3 \\ 
ABMIL & 77.8 & 50.6 & 40.4 & 75.6 & 37.0 & 27.4 \\ 
MHIM-ABMIL & 78.4 & 51.2 & 41.5 & 76.1 & 37.0 & 28.7 \\ 
DSMIL & 78.7 & 46.7 & 35.6 & 74.9 & 37.0 & 29.9 \\ 
DTFD-MIL & 80.9 & 49.2 & 39.2 & 76.3 & 35.0 & 27.4 \\ 
TransMIL & 73.1 & 42.1 & 24.8 & 71.4 & 37.0 & 27.6 \\ 
MHIM-TransMIL & 74.5 & 44.5 & 28.5 & 72.5 & 37.0 & 28.5 \\ 
S4MIL & 77.4 & 46.5 & 36.2 & 72.5 & 33.0 & 22.5 \\ 
RRTMIL & 79.5 & 48.5 & 37.9 & 76.8 & 34.0 & 22.5 \\ 
MambaMIL & 78.0 & 47.0 & 32.6 & 76.6 & 35.0 & 27.7 \\ 
CAMIL & 79.7 & 49.8 & 37.5 & 75.2 & 37.0 & 27.0 \\ 
RPMIL & 81.7 & 51.6 & 42.5 & 77.5 & 38.0 & 31.4 \\ 
ACMIL & 81.2 & 51.0 & 41.6 & 77.2 & 38.0 & 30.7 \\ 
Jigsaw & 82.0 & 52.0 & 43.0 & 77.6 & 38.0 & 31.8 \\ 
LEAF-Mamba & \textbf{83.2} & \textbf{53.6} & \textbf{44.9} & \textbf{78.6} & \textbf{40.0} & \textbf{32.7} \\ 
\bottomrule
\end{tabular*}
\par\smallskip{\footnotesize Best scores in each column are in bold.\par}
\end{minipage} \\
\noalign{\vskip 10pt}
\begin{minipage}{\linewidth}
\fontsize{10}{12}\selectfont
\textbf{(b) Camelyon16 and EBRAINS}\par\smallskip
\begin{tabular*}{\linewidth}{@{\extracolsep{\fill}}lrrrrrr@{}}
\toprule
& \multicolumn{3}{c}{Camelyon16} & \multicolumn{3}{c}{EBRAINS} \\
\cmidrule(lr){2-4}\cmidrule(lr){5-7}
Method & AUC & ACC & F1 & AUC & ACC & F1 \\
\midrule
MeanMIL & 91.5 & 93.9 & 88.3 & 97.2 & 63.6 & 62.6 \\ 
MaxMIL & 87.7 & 90.2 & 89.7 & 97.4 & 67.6 & 67.0 \\ 
CLAM & 94.8 & 92.7 & 92.3 & 97.8 & 66.9 & 65.9 \\ 
ABMIL & 96.5 & 93.2 & 91.5 & 97.7 & 67.3 & 66.4 \\ 
MHIM-ABMIL & 96.9 & 93.4 & 91.5 & \textbf{97.9} & 67.9 & 67.1 \\ 
DSMIL & 92.7 & 88.5 & 84.8 & 97.8 & 66.7 & 65.9 \\ 
DTFD-MIL & 97.5 & \textbf{94.2} & \textbf{92.6} & 97.6 & 67.9 & 67.0 \\ 
TransMIL & 96.5 & 90.5 & 88.8 & 97.3 & 60.4 & 58.4 \\ 
MHIM-TransMIL & 96.9 & 91.7 & 89.7 & 97.6 & 61.8 & 59.7 \\ 
S4MIL & 95.8 & 94.0 & 92.5 & 97.5 & 63.2 & 62.5 \\ 
RRTMIL & 96.2 & 92.5 & 92.1 & 97.2 & 63.8 & 62.9 \\ 
MambaMIL & 97.0 & 92.5 & 90.7 & 97.2 & 64.3 & 62.6 \\ 
CAMIL & 96.2 & 92.9 & 91.8 & 96.1 & 62.7 & 64.5 \\ 
RPMIL & 97.8 & 94.0 & 92.4 & 95.8 & 64.2 & 65.7 \\ 
ACMIL & 97.5 & 93.8 & 92.2 & 96.1 & 64.4 & 64.3 \\ 
Jigsaw & 97.8 & 94.0 & 92.5 & 96.2 & 64.8 & 66.3 \\ 
LEAF-Mamba & \textbf{97.9} & \textbf{94.2} & 92.5 & \textbf{97.9} & \textbf{68.2} & \textbf{68.5} \\ 
\bottomrule
\end{tabular*}
\par\smallskip{\footnotesize Best scores in each column are in bold.\par}
\end{minipage} \\
\noalign{\vskip 10pt}
\begin{minipage}{\linewidth}
\fontsize{10}{12}\selectfont
\textbf{(c) Average}\par\smallskip
\begin{tabular*}{\linewidth}{@{\extracolsep{\fill}}lrrr@{}}
\toprule
Method & AUC & ACC & F1 \\
\midrule
MeanMIL & 82.5 & 59.7 & 52.1 \\ 
MaxMIL & 83.4 & 59.7 & 54.9 \\ 
CLAM & 86.6 & 61.6 & 55.0 \\ 
ABMIL & 86.9 & 62.0 & 56.4 \\ 
MHIM-ABMIL & 87.3 & 62.4 & 57.2 \\ 
DSMIL & 86.0 & 59.7 & 54.1 \\ 
DTFD-MIL & 88.1 & 61.6 & 56.6 \\ 
TransMIL & 84.5 & 57.5 & 49.9 \\ 
MHIM-TransMIL & 85.4 & 58.8 & 51.6 \\ 
S4MIL & 85.8 & 59.2 & 53.4 \\ 
RRTMIL & 87.4 & 59.7 & 53.9 \\ 
MambaMIL & 87.2 & 59.7 & 53.4 \\ 
CAMIL & 86.8 & 60.6 & 55.2 \\ 
RPMIL & 88.2 & 62.0 & 58.0 \\ 
ACMIL & 88.0 & 61.8 & 57.2 \\ 
Jigsaw & 88.4 & 62.2 & 58.4 \\ 
LEAF-Mamba & \textbf{89.4} & \textbf{64.0} & \textbf{59.6} \\ 
\bottomrule
\end{tabular*}
\par\smallskip{\footnotesize Best scores in each column are in bold. Average summarizes the four diagnostic settings.\par}
\end{minipage} \\
\end{longtable}
\endgroup

\begingroup
\setstretch{1}
\fontsize{10}{12}\selectfont
\setlength{\tabcolsep}{3pt}
\renewcommand{\arraystretch}{1.0}
\setlength{\LTcapwidth}{\linewidth}
\setlength{\LTpre}{10pt}
\setlength{\LTpost}{12pt}
\begin{longtable}{@{}p{\linewidth}@{}}
\caption{Survival prediction results averaged over five TCGA cohorts. Values are C-index $\times 100$; higher is better.}\label{tab:survival}\\
\endfirsthead
\caption[]{Survival prediction results averaged over five TCGA cohorts. Values are C-index $\times 100$; higher is better (continued).}\\
\endhead
\endfoot
\endlastfoot
\begin{minipage}{\linewidth}
\fontsize{10}{12}\selectfont
\textbf{Five-cohort average}\par\smallskip
\begin{tabular*}{\linewidth}{@{\extracolsep{\fill}}lrrr@{}}
\toprule
Method & ResNet-50 & PLIP & CONCH v1.5 \\
\midrule
CLAM & 58.8 & 59.0 & 64.4 \\ 
ABMIL & 59.1 & 60.9 & 62.5 \\ 
TransMIL & 65.5 & 64.9 & 63.9 \\ 
S4MIL & 65.1 & 63.0 & 64.7 \\ 
RRTMIL & 66.0 & 63.7 & 65.2 \\ 
MambaMIL & 66.7 & 66.0 & 65.5 \\ 
CAMIL & 65.7 & 65.2 & 64.6 \\ 
RPMIL & 66.1 & 66.2 & 65.6 \\ 
ACMIL & 66.0 & 65.7 & 65.0 \\ 
Jigsaw & 66.9 & 67.1 & 67.0 \\ 
LEAF-Mamba & \textbf{67.0} & \textbf{67.3} & \textbf{67.3} \\ 
\bottomrule
\end{tabular*}
\par\smallskip{\footnotesize Best scores in each column are in bold.\par}
\end{minipage} \\
\end{longtable}
\endgroup

While Figure~\ref{fig:main_results} establishes the overall predictive performance, Tables~\ref{tab:ablation_fusion}--\ref{tab:ablation_components} examine the contributions of the adaptation module, sequence branches, and fusion mechanism under controlled configurations.

\subsection{Ablation Study}

Tables~\ref{tab:ablation_fusion}--\ref{tab:ablation_components} evaluate three aspects of \method{}: branch and fusion design, latent feature adaptation, and component behavior across feature extractors and prediction tasks. The branch/fusion and adapter comparisons use the BRACS official split with CONCH v1.5 features, while the cross-setting analysis additionally includes PLIP features, Camelyon16, and TCGA-BRCA.

\FloatBarrier

\begingroup
\setstretch{1}
\fontsize{10}{12}\selectfont
\setlength{\tabcolsep}{3pt}
\renewcommand{\arraystretch}{1.0}
\setlength{\LTcapwidth}{\linewidth}
\setlength{\LTpre}{10pt}
\setlength{\LTpost}{12pt}
\begin{longtable}{@{}p{\linewidth}@{}}
\caption{Branch and fusion comparisons on BRACS official with CONCH v1.5 features. Values are percentages; higher is better.}\label{tab:ablation_fusion}\\
\endfirsthead
\caption[]{Branch and fusion comparisons on BRACS official with CONCH v1.5 features. Values are percentages; higher is better (continued).}\\
\endhead
\endfoot
\endlastfoot
\begin{minipage}{\linewidth}
\fontsize{10}{12}\selectfont
\textbf{Branch and fusion ablation}\par\smallskip
\begin{tabular*}{\linewidth}{@{\extracolsep{\fill}}l l >{\raggedright\arraybackslash}p{58pt} >{\raggedright\arraybackslash}p{48pt} r r r@{}}
\toprule
ID & LEA & Branches & Fusion & AUC & ACC & F1 \\
\midrule
F1 & Yes & Local only & -- & 75.7 & 35.3 & 27.4 \\ 
F2 & Yes & Global only & -- & 76.8 & 36.4 & 29.2 \\ 
F3 & Yes & Global + local & Concat & 77.2 & 37.0 & 30.4 \\ 
F4 & Yes & Global + local & Add & 77.0 & 36.9 & 31.0 \\ 
F5 & Yes & Global + local & Mean & 77.1 & 36.8 & 30.8 \\ 
F6 & Yes & Global + local & Equal gate & 77.5 & 37.6 & 31.4 \\ 
F7 & Yes & Global + local & Static gate & 77.8 & 37.9 & 31.3 \\ 
F8 & Yes & Global + local & Gated & \textbf{78.6} & \textbf{40.0} & \textbf{32.7} \\ 
F9 & -- & Global + local & Concat & 76.7 & 37.0 & 29.8 \\ 
\bottomrule
\end{tabular*}
\par\smallskip{\footnotesize LEA denotes latent evidence adaptation. F8 is the complete model. Best scores in each metric are in bold.\par}
\end{minipage} \\
\end{longtable}
\endgroup

\begingroup
\setstretch{1}
\fontsize{10}{12}\selectfont
\setlength{\tabcolsep}{3pt}
\renewcommand{\arraystretch}{1.0}
\setlength{\LTcapwidth}{\linewidth}
\setlength{\LTpre}{10pt}
\setlength{\LTpost}{12pt}
\begin{longtable}{@{}p{\linewidth}@{}}
\caption{Adapter comparisons on BRACS official with CONCH v1.5 features under global-only aggregation. Performance values are percentages.}\label{tab:ablation_adapter}\\
\endfirsthead
\caption[]{Adapter comparisons on BRACS official with CONCH v1.5 features under global-only aggregation. Performance values are percentages (continued).}\\
\endhead
\endfoot
\endlastfoot
\begin{minipage}{\linewidth}
\fontsize{10}{12}\selectfont
\textbf{Adapter performance and parameter counts}\par\smallskip
\begin{tabular*}{\linewidth}{@{\extracolsep{\fill}}>{\raggedright\arraybackslash}p{96pt} r r r r r@{}}
\toprule
& \multicolumn{2}{c}{Parameters (K)} & & & \\
\cmidrule(lr){2-3}
Adapter & Train & Infer. & AUC & ACC & F1 \\
\midrule
No adapter (frozen PCA bridge) & 0.0 & 0.0 & 75.1 & 34.4 & 27.4 \\ 
Linear projection & 196.9 & 196.9 & 75.5 & \textbf{36.4} & 27.9 \\ 
Two-layer MLP & 721.9 & 721.9 & \textbf{76.8} & 36.1 & 28.3 \\ 
Residual bottleneck adapter & 722.4 & 722.4 & 76.4 & 35.7 & 28.5 \\ 
LEA without reconstruction & 723.5 & 723.5 & 75.8 & 35.8 & 28.1 \\ 
Full LEA & 1250.0 & 723.5 & \textbf{76.8} & \textbf{36.4} & \textbf{29.2} \\ 
\bottomrule
\end{tabular*}
\par\smallskip{\footnotesize K denotes 1,000 parameters; counts are rounded to one decimal place. Training includes the reconstruction decoder; inference excludes it. Best performance scores are in bold.\par}
\end{minipage} \\
\end{longtable}
\endgroup

\begingroup
\setstretch{1}
\fontsize{10}{12}\selectfont
\setlength{\tabcolsep}{3pt}
\renewcommand{\arraystretch}{1.0}
\setlength{\LTcapwidth}{\linewidth}
\setlength{\LTpre}{10pt}
\setlength{\LTpost}{12pt}
\begin{longtable}{@{}p{\linewidth}@{}}
\caption{Cross-setting component comparisons. Classification values are percentages; survival values are C-index $\times 100$. Higher is better.}\label{tab:ablation_components}\\
\endfirsthead
\caption[]{Cross-setting component comparisons. Classification values are percentages; survival values are C-index $\times 100$. Higher is better (continued).}\\
\endhead
\endfoot
\endlastfoot
\begin{minipage}{\linewidth}
\fontsize{10}{12}\selectfont
\textbf{(a) BRACS official / CONCH v1.5}\par\smallskip
\begin{tabular*}{\linewidth}{@{\extracolsep{\fill}}l l >{\raggedright\arraybackslash}p{101pt} rrr@{}}
\toprule
ID & LEA & Aggregation & AUC & ACC & F1 \\
\midrule
C1 & -- & Global only & 76.6 & 35.3 & 27.7 \\ 
C2 & Yes & Global only & 76.8 & 36.4 & 29.2 \\ 
C3 & -- & Global + local$^{*}$ & 77.2 & 37.4 & 31.6 \\ 
C4 & Yes & Global + local$^{*}$ & \textbf{78.6} & \textbf{40.0} & \textbf{32.7} \\ 
\bottomrule
\end{tabular*}
\end{minipage} \\
\noalign{\vskip 10pt}
\begin{minipage}{\linewidth}
\fontsize{10}{12}\selectfont
\textbf{(b) BRACS official / PLIP}\par\smallskip
\begin{tabular*}{\linewidth}{@{\extracolsep{\fill}}l l >{\raggedright\arraybackslash}p{101pt} rrr@{}}
\toprule
ID & LEA & Aggregation & AUC & ACC & F1 \\
\midrule
C1 & -- & Global only & 76.9 & 35.9 & 26.5 \\ 
C2 & Yes & Global only & 75.5 & 34.4 & 27.6 \\ 
C3 & -- & Global + local$^{*}$ & 77.2 & 29.6 & 29.3 \\ 
C4 & Yes & Global + local$^{*}$ & \textbf{77.8} & \textbf{36.0} & \textbf{29.8} \\ 
\bottomrule
\end{tabular*}
\end{minipage} \\
\noalign{\vskip 10pt}
\begin{minipage}{\linewidth}
\fontsize{10}{12}\selectfont
\textbf{(c) Camelyon16 / CONCH v1.5}\par\smallskip
\begin{tabular*}{\linewidth}{@{\extracolsep{\fill}}l l >{\raggedright\arraybackslash}p{101pt} rrr@{}}
\toprule
ID & LEA & Aggregation & AUC & ACC & F1 \\
\midrule
C1 & -- & Global only & 97.2 & 92.2 & 89.9 \\ 
C2 & Yes & Global only & 97.8 & 92.7 & 89.7 \\ 
C3 & -- & Global + local$^{*}$ & 97.5 & 93.8 & \textbf{92.7} \\ 
C4 & Yes & Global + local$^{*}$ & \textbf{97.9} & \textbf{94.2} & 92.5 \\ 
\bottomrule
\end{tabular*}
\end{minipage} \\
\noalign{\vskip 10pt}
\begin{minipage}{\linewidth}
\fontsize{10}{12}\selectfont
\textbf{(d) TCGA-BRCA / CONCH v1.5}\par\smallskip
\begin{tabular*}{\linewidth}{@{\extracolsep{\fill}}l l >{\raggedright\arraybackslash}p{101pt} r@{}}
\toprule
ID & LEA & Aggregation & C-index \\
\midrule
C1 & -- & Global only & 67.2 \\ 
C2 & Yes & Global only & \textbf{68.1} \\ 
C3 & -- & Global + local$^{*}$ & 66.0 \\ 
C4 & Yes & Global + local$^{*}$ & 68.0 \\ 
\bottomrule
\end{tabular*}
\par\smallskip{\footnotesize $^{*}$ denotes gated global--local fusion. C4 is the complete model. Best scores in each setting are in bold.\par}
\end{minipage} \\
\end{longtable}
\endgroup

Table~\ref{tab:ablation_fusion} first isolates the aggregation design with \lsr{} enabled. The global-only model achieves 76.8 AUC, 36.4 ACC, and 29.2 F1, exceeding the local-only model on all three metrics. Combining both branches through concatenation, addition, or averaging further improves all three scores over global-only aggregation. Input-dependent gated fusion achieves the highest scores of 78.6 AUC, 40.0 ACC, and 32.7 F1, improving on the global-only model by 1.8, 3.6, and 3.5 percentage points, respectively. Compared with the learned static gate, it gains 0.8 AUC, 2.1 ACC, and 1.4 F1 points. These comparisons support both the use of parallel long-range and sequence-local branches and input-dependent modulation of their joint representation.

Table~\ref{tab:ablation_adapter} compares adapters under global-only aggregation. Full \lsr{} achieves 76.8 AUC, 36.4 ACC, and 29.2 F1, compared with 75.1, 34.4, and 27.4 for the frozen PCA bridge. It matches the two-layer MLP in AUC while improving ACC and F1 by 0.3 and 0.9 points. Removing the reconstruction objective reduces AUC, ACC, and F1 to 75.8, 35.8, and 28.1, respectively, demonstrating the contribution of reconstruction regularization in this setting. Full \lsr{} uses 1,250.0K adapter parameters during training and 723.5K during inference; the reconstruction decoder adds training capacity while leaving the inference parameter count equal to that of \lsr{} without reconstruction.

Table~\ref{tab:ablation_components} examines how adaptation and aggregation work together across four evaluation settings. Relative to global-only aggregation without \lsr{}, the complete model raises AUC from 76.6 to 78.6 on BRACS with CONCH, from 76.9 to 77.8 on BRACS with PLIP, and from 97.2 to 97.9 on Camelyon16. It achieves the highest AUC and ACC in all three classification settings, and the highest F1 in both BRACS settings. On Camelyon16, gated global--local aggregation without \lsr{} has a slightly higher F1 of 92.7 versus 92.5 for the complete model. The effect of adaptation also depends on the aggregator: with PLIP features, adding \lsr{} changes global-only AUC from 76.9 to 75.5, but improves gated global--local AUC from 77.2 to 77.8. For TCGA-BRCA, \lsr{} raises the global-only C-Index from 67.2 to 68.1, while the complete model scores 68.0. Thus, the combined design delivers the strongest classification AUC across the tested settings, while the survival comparison attributes the main improvement to feature adaptation.

\subsection{Qualitative Case Study}

\begin{figure}[!t]
\centering
\includegraphics[width=\textwidth]{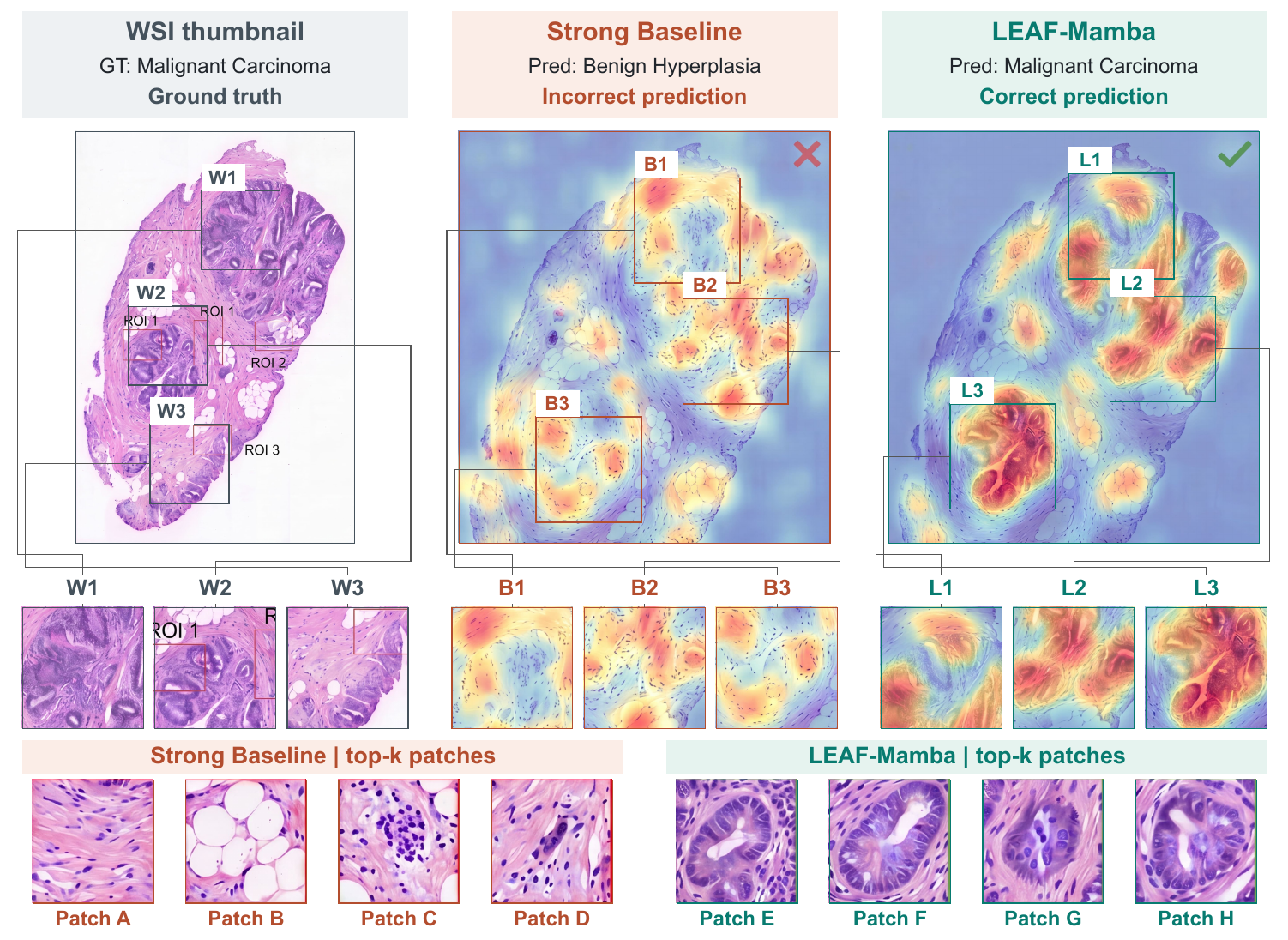}
\caption{Qualitative attention comparison on a BRACS malignant carcinoma case. The displayed top-$k$ patches differ between the models, and \method{} predicts the correct malignant carcinoma label in this example.}
\label{fig:case_study}
\end{figure}

Figure~\ref{fig:case_study} shows attention patterns on one BRACS malignant carcinoma example. The baseline predicts benign hyperplasia, whereas \method{} predicts malignant carcinoma; the displayed top-$k$ patches and marked regions provide a qualitative comparison. Work on attention explanations shows that interpretation depends on the model, the definition of explanation, and appropriate counterfactual tests~\cite{jain2019attention,wiegreffe2019attention}. We consequently use this case to describe the model's selection pattern, without inferring localization accuracy, faithful attribution, or clinical utility from one heatmap.

\FloatBarrier
\section{Discussion and Limitations}
\label{sec:discussion}

\subsection{Interpretation of the Component Analyses}

The component analyses distinguish the effects of feature adaptation from those of context fusion.
In the BRACS branch comparison, parallel global--local processing improves on either branch alone, and input-dependent fusion further improves on a learned static gate.
In the adapter comparison, removing reconstruction reduces performance, supporting the use of input recovery as an auxiliary training objective.
These results connect the design choices to their controlled variants: adaptation changes the features available to aggregation, while fusion changes how complementary sequence representations are combined.

The cross-setting results also show that the contributions of the two components depend on the task and the aggregator.
With PLIP features on BRACS, adaptation improves the gated global--local model but does not improve the global-only model's AUC.
For TCGA-BRCA, the adapted global-only variant slightly exceeds the complete model in C-Index.
Thus, the strongest average benchmark performance of the complete framework does not imply that each component improves every setting.
The evidence supports evaluating adaptation and aggregation jointly, with their individual contributions assessed under matched feature extractors and protocols.

\subsection{Scope and Limitations}
Our study focuses on learning from frozen patch representations, enabling controlled comparisons of adaptation and aggregation across feature extractors. Within this setting, the information available to the model depends on the pretrained encoder; exploring selective encoder fine-tuning would help establish when adaptation at the feature level is sufficient and when updating the encoder offers additional benefits. In addition, local interactions are defined along a fixed coordinate-derived sequence. Comparing alternative spatial orderings and incorporating explicit two-dimensional relationships are promising directions for extending the fusion design. Finally, the current evaluation covers multiple public benchmarks, while transfer across institutions remains an important next step. Independent cohorts with different patient populations and slide preparation practices would help characterize the conditions under which the observed benefits generalize.

\section{Conclusion}
\label{sec:conclusion}

We presented \method{}, a frozen-feature WSI framework combining reconstruction-regularized token adaptation, long-range sequence modeling, sequence-local convolution, and input-dependent fusion. Our aim was to address a central challenge in weakly supervised computational pathology: adapting pretrained patch representations to downstream tasks while bringing together complementary local and global information. Across feature extractors and prediction tasks, \method{} achieves strong overall performance in both diagnostic classification and survival prediction. The component analyses further highlight the value of jointly designing feature adaptation and contextual aggregation. More broadly, these findings suggest that effective use of pathology foundation models depends on how their representations are adapted and integrated into slide-level decisions. This perspective offers a direction for connecting reusable pretrained features with diverse pathology tasks. Future work will extend this approach to cross-institutional evaluation and multimodal learning, exploring how tissue representations can be combined with complementary patient information for more comprehensive disease characterization.

\bibliographystyle{elsarticle-num}
\bibliography{custom}

\end{document}